\documentclass{article} 
\usepackage{iclr2017_conference,times}
\usepackage{hyperref}
\usepackage{url}
\usepackage{multirow}

\usepackage{framed}
\usepackage{algorithmic}
\usepackage{graphicx} 
\usepackage{subfigure}
\usepackage{amssymb}
\usepackage{amsmath}
\usepackage{amsthm}
\usepackage{caption}
\usepackage{tabularx}
\usepackage{listings}
\usepackage{url}
\usepackage{verbatimbox}
\usepackage[export]{adjustbox}

\newcommand{\script}[2]{
}

\usepackage[compact]{titlesec}
\titlespacing{\section}{0pt}{0.5ex}{0.3ex}
\titlespacing{\subsection}{0pt}{0.2ex}{0ex}
\titlespacing{\subsubsection}{0pt}{0.1ex}{0ex}

\usepackage{paralist}

\makeatletter
\ifcase \@ptsize \relax
  \newcommand{\miniscule}{\@setfontsize\miniscule{4}{5}}
\or
  \newcommand{\miniscule}{\@setfontsize\miniscule{5}{6}}
\or
  \newcommand{\miniscule}{\@setfontsize\miniscule{5}{6}}
\fi
\makeatother

\newcommand {\aplt} {\ {\raise-.5ex\hbox{$\buildrel<\over\sim$}}\ }

\newcommand{\pnoise}{p_{\text{Noise}}}

\newcommand{\fig}[1]{Fig.~\ref{fig:#1}}
\newcommand{\tab}[1]{Table~\ref{tab:#1}}

\newcommand {\R}{\mathbb {R}}

\title{Semi-Supervised Learning with \\Context-Conditional Generative \\Adversarial Networks}

\author{Emily Denton\\
Dept. of Computer Science\\
Courant Institute, New York University\\
\texttt{denton@cs.nyu.edu}
\And
Sam Gross\\
Facebook AI Research \\
New York\\
\texttt{sgross@fb.com}
\And
Rob Fergus \\
Facebook AI Research \\
New York\\
\texttt{robfergus@fb.com}
}

\begin{document}

\maketitle  

\vspace{-5mm}
\begin{abstract}
  We introduce a simple semi-supervised learning approach for images based on
  in-painting using an adversarial loss. Images with random
  patches removed are presented to a generator whose task is to fill
  in the hole, based on the surrounding pixels. The in-painted images
  are then presented to a discriminator network that judges if they
  are real (unaltered training images) or not. This task acts as a
  regularizer for standard supervised training of the
  discriminator. Using our approach we are able to directly train
  large VGG-style networks in a semi-supervised fashion. We evaluate
  on STL-10 and PASCAL datasets, where our approach obtains
  performance comparable or superior to existing methods.
\end{abstract}

\section{Introduction}

Deep neural networks have yielded dramatic performance gains in recent
years on tasks such as object classification \citep{krizhevsky2012},
text classification \citep{zhang2015} and machine translation
\citep{sutskever2014,bahdanau2015}.  These successes are heavily
dependent on large training sets of manually annotated data.  In many
settings however, such large collections of labels may not be readily
available, motivating the need for methods that can learn from data
where labels are rare.

We propose a method for harnessing unlabeled image data based on
image in-painting. A generative model is trained to generate pixels within a
missing hole, based on the context provided by surrounding parts of
the image. These in-painted images are then used in an adversarial
setting \citep{goodfellow2014} to train a large discriminator model whose task is
to determine if the image was real (from the unlabeled training
set) or fake (an in-painted image). The realistic looking
fake examples provided by the generative model cause the discriminator
to learn features that generalize to the related task of classifying
objects. Thus adversarial training for the in-painting task can be
used to regularize large discriminative models during supervised
training on a handful of labeled images.


\subsection{Related Work}

\begin{figure}[t]
\centering
   \includegraphics[width=1\linewidth]{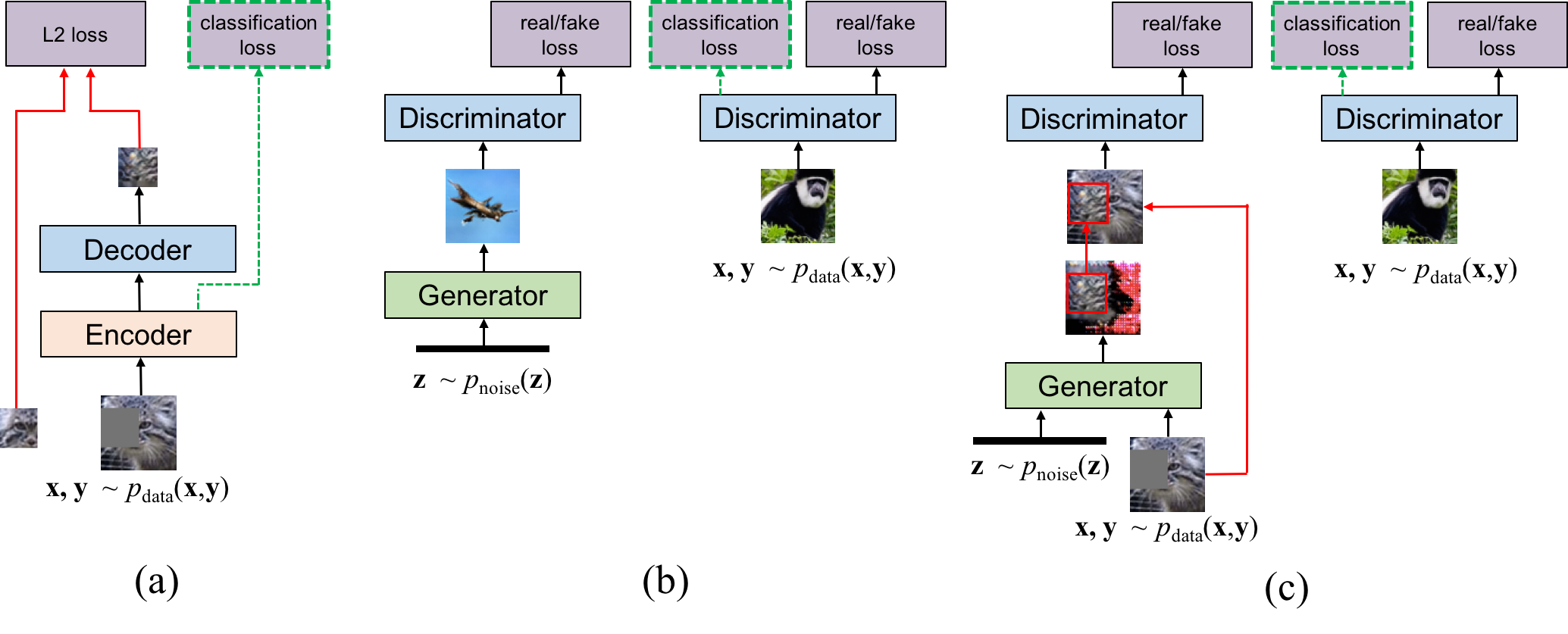} 
 \caption{ {\bf(a)} Context-encoder of \cite{pathak2016}, configured for object classification task. 
   {\bf(b)} Semi-supervised learning with GANs (SSL-GAN). 
   {\bf(c)} Semi-supervised learning with CC-GANs. In {\bf (a-c)} the blue network indicates the feature representation being learned (encoder network in the context-encoder model and discriminator network in the GAN and CC-GAN models).
}
 \label{fig:semisup}
\end{figure}


\textbf{Learning From Context:} 
The closest work to ours is the independently
developed context-encoder approach of 
\cite{pathak2016}.
This introduces an encoder-decoder framework, shown in
\fig{semisup}(a), that is used to in-paint images where a patch has
been randomly removed. After using this as a pre-training task, a
classifier is added to the encoder and the model is fine-tuned using
the labeled examples. Although both approaches use the concept of
in-painting, they differ in several important ways. First, the
architectures are different (see \fig{semisup}): in \cite{pathak2016},
the features for the classifier are taken from the encoder, whereas
ours come from the discriminator network. In practice this makes an
important difference as we are able to directly train large models
such as VGG \citep{vgg} using adversarial loss alone. By contrast,
\cite{pathak2016} report difficulties in training an
AlexNet encoder with this loss. This leads to the second difference,
namely that on account of these issues, they instead employ an
$\ell_2$ loss when training models for classification and detection
(however they do use a joint $\ell_2$ and adversarial loss to achieve
impressive in-painting results).  Finally, the unsupervised learning
task differs between the two models.  The context-encoder learns a
feature representation suitable for in-painting whereas our model
learns a feature representation suitable for differentiating real/fake
in-paintings.  Notably, while we also use a neural network to generate
the in-paintings, this model is only used as an adversary for the
discriminator, rather than as a feature extractor.  In section 4, we
compare the performance of our model to the context-encoder on the
PASCAL dataset.


Other forms of spatial context within images have recently been
utilized for representation learning.  
\cite{doersch2015} propose training a CNN to predict the spatial
location of one image patch relative to another.  
\cite{noroozi2016} propose a model that learns by unscrambling image
patches, essentially solving a jigsaw puzzle to learn visual
representations. In the text domain, context has been successfully
leveraged as an unsupervised criterion for training useful word and
sentence level representations \citep{collobert2011, mikolov2013,
  kiros2015}.  

\textbf{Deep unsupervised and semi-supervised learning:} A popular method
of utilizing unlabeled data is to layer-wise train a deep autoencoder
or restricted Botlzmann machine \citep{hinton2006} and then fine tune
with labels on a discriminative task.  More recently, several
autoencoding variants have been proposed for unsupervised and
semi-supervised learning, such as the ladder network \citep{ramus2015},
stacked what-where autoencoders \citep{zhao2016} and variational
autoencoders \citep{kingma2014a, kingma2014b}.

\cite{dosovitskiy2014b} achieved state-of-the-art results
by training a CNN with a different class for each training example and
introducing a set of transformations to provide multiple examples per
class.  The pseudo-label approach
\citep{lee2013} is a simple semi-supervised method that trains using
the maximumly predicted class as a label when labels are unavailable.
\cite{Springenberg2015} propose a categorical generative adversarial network (CatGAN) which can be used for unsupervised and semi-supervised learning. 
The discriminator in a CatGAN outputs a distribution over classes and is trained to minimize the predicted entropy for real data and maximize the predicted entropy for fake data. 
Similar to our model, CatGANs use the feature space learned by the discriminator for the final supervised learning task. 
\cite{Salimans2016} recently proposed a semi-supervised GAN model in which the discriminator outputs a softmax over classes rather than a probability of real vs. fake. 
An additional `generated' class is used as the target for generated samples. 
This method differs from our work in that it does not utilize context information and has only been applied to datasets of small resolution. 
However, the discriminator loss is similar to the one we propose and could be combined with our context-conditional approach.

More traditional semi-supervised methods include graph-based
approaches \citep{Zhuo04,Zhu06semi-supervisedlearning} that show
impressive performance when good image representations are
available. However, the focus of our work is on learning such
representations.

\textbf{Generative models of images:} Restricted Boltzmann machines \citep{salakhutdinov2015},
de-noising autoencoders \citep{vincent2008} and variational
autoencoders \citep{kingma2014a} optimize a maximum likelihood
criterion and thus learn decoders that map from latent space to image
space.  More recently, generative adversarial networks
\citep{goodfellow2014} and generative moment matching networks
\citep{li2015,karolina2015} have been proposed.  These methods ignore data likelihoods
and instead directly train a generative model to produce realistic
samples.  Several extensions to the generative adversarial network
framework have been proposed to scale the approach to larger images
\citep{denton2015,radford2016,Salimans2016}. Our
work draws on the insights of \cite{radford2016} regarding adversarial training practices
and architecture for the generator network, as well as the notion that
the discriminator can produce useful features for classification
tasks.

Other models used recurrent approaches to generate images
\citep{gregor2015, theis2015, mansimov2016, oord2016}.  \cite{dosovitskiy2015} trained a CNN to generate objects with
different shapes, viewpoints and color.  
\cite{sohldickstein2015} propose a generative model based on a reverse
diffusion process. While our model does involve image generation, it
differs from these approaches in that the main focus is on learning a
good representation for classification tasks.

Predictive generative models of videos aim to extrapolate from current
frames to future ones and in doing so learn a feature
representation that is useful for other tasks. In this vein, \cite{Ranzato14} used an
$\ell_2$-loss in pixel-space. \cite{Mathieu15} combined
an adversarial loss with $\ell_2$, giving models that generate crisper
images. While our model is also predictive, it only
considers interpolation within an image, rather than extrapolation in
time.

\section{Approach}
We present a semi-supervised learning framework built on generative adversarial networks (GANs) of \cite{goodfellow2014}. 
We first review the generative adversarial network framework and then introduce context conditional generative adversarial networks (CC-GANs).
Finally, we show how combining a classification objective and a CC-GAN objective provides a unified framework for semi-supervised learning.

\subsection{Generative Adversarial Networks}
The generative adversarial network approach \citep{goodfellow2014} is a
framework for training generative models, which we briefly review.  
It consists of two networks pitted against one another in a two player game:
A generative model, $G$, is trained to synthesize images resembling the data distribution and a discriminative model, $D$, is trained to distinguish between samples drawn from $G$ and images drawn from the training data.

More formally, let $\mathcal{X} = \{\mathbf{x}^1, ..., \mathbf{x}^n\}$ be a dataset of images of dimensionality $d$.
Let $D$ denote a discriminative function that takes as input an image $\mathbf{x} \in \R^d$ and outputs a scalar representing the probability  of input $\mathbf{x}$ being a real sample.
Let $G$ denote the generative function that takes as input a random vector $\mathbf{z} \in \R^z$ sampled from a prior noise distribution $\pnoise$ and outputs a synthesized image $\mathbf{\tilde{x}} = G(\mathbf{z}) \in \R^d$. Ideally, $D(\mathbf{x}) = 1$ when $\mathbf{x} \in \mathcal{X}$ and $D(\mathbf{x}) = 0$ when $\mathbf{x}$ was generated from $G$.
The GAN objective is given by:
\begin{equation}
\label{eqn:gan}
\min_G \max_D \hspace{2mm} \mathbb{E}_{\mathbf{x}\sim \mathcal{X}}[\log D(\mathbf{x})]  +
\mathbb{E}_{\mathbf{z} \sim \pnoise(\mathbf{z})} [\log(1 - D(G(\mathbf{z})))]
\end{equation}

The conditional generative adversarial network \citep{mirza14} is an extension of the GAN in which both $D$ and $G$ receive an additional vector of information $\mathbf{y}$ as input. The conditional GAN objective is given by:
\begin{equation}
\label{eqn:cgan}
\min_G \max_D \hspace{2mm} \mathbb{E}_{\mathbf{x, y}\sim \mathcal{X}}[\log D(\mathbf{x, y})]  +
\mathbb{E}_{\mathbf{z} \sim \pnoise(\mathbf{z})} [\log(1 - D(G(\mathbf{z, y}), \mathbf{x}))]
\end{equation}

\subsection{Context-Conditional Generative Adversarial Networks}
We propose context-conditional generative adversarial networks (CC-GANs) which are
conditional GANs where the generator is trained to
fill in a missing image patch and the generator and discriminator are
conditioned on the surrounding pixels. 

In particular, the generator $G$ receives as input an image with a
randomly masked out patch.  The generator outputs an entire image. 
We fill in the missing patch from the generated output and then pass the completed image
into $D$.  
We pass the completed image into $D$ rather than the context and the patch as two separate inputs so as to prevent $D$ from simply learning to identify discontinuities along the edge of the missing patch.

More formally, let $\mathbf{m} \in \mathbb{R}^d$ denote to a binary mask that will be used to drop out a specified portion of an image. 
The generator receives as input $\mathbf{m} \odot \mathbf{x}$ where $\odot$ denotes element-wise multiplication.
The generator outputs $\mathbf{x_G} = G(\mathbf{m} \odot \mathbf{x}, \mathbf{z})
\in \mathbb{R}^d$ and the in-painted image $\mathbf{x_I}$ is given by: 
\begin{equation}
\mathbf{x_I} = (1 - \mathbf{m}) \odot \mathbf{x_G} + \mathbf{m} \odot \mathbf{x}
\end{equation}

The CC-GAN objective is given by:
\begin{equation}
\label{eqn:ccgan}
\min_G \max_D \hspace{2mm}\mathbb{E}_{\mathbf{x} \sim \mathcal{X}}[\log D(\mathbf{x})]  +
\mathbb{E}_{\mathbf{x} \sim \mathcal{X}, \mathbf{m} \sim \mathcal{M}} [\log(1 - D(\mathbf{x_I}))]
\end{equation}

 
\subsection{Combined GAN and CC-GAN}
While the generator of the CC-GAN outputs a full image, only a portion of it (corresponding to the missing hole) is seen by the discriminator. 
In the combined model, which we denote by CC-GAN$^2$, the fake
examples for the discriminator include both the in-painted image $\mathbf{x_I}$ and
the full image $\mathbf{x_G}$ produced by the generator (i.e. not just the missing patch). 
By combining the GAN and CC-GAN approaches, we introduce a wider array of negative examples to the discriminator. The CC-GAN$^2$ objective given by:
\begin{align}
\label{eqn:ccgan_comb}
\min_G \max_D \hspace{2mm} & \mathbb{E}_{\mathbf{x} \sim \mathcal{X}}[\log D(\mathbf{x})]  \\
+ \hspace{2mm} &\mathbb{E}_{\mathbf{x} \sim \mathcal{X}, \mathbf{m} \sim \mathcal{M}} [\log(1 - D(\mathbf{x_I}))]\\
+ \hspace{2mm} &\mathbb{E}_{\mathbf{x} \sim \mathcal{X}, \mathbf{m} \sim \mathcal{M}} [\log(1 - D(\mathbf{x_G} ))]
\end{align}

\subsection{Semi-supervised learning with CC-GANs}
A common approach to semi-supervised learning is to combine a supervised and unsupervised objective during training.
As a result unlabeled data can be leveraged to aid the supervised task.

Intuitively, a GAN discriminator must learn something about the structure of natural images in order to effectively distinguish real from generated images. 
Recently, \cite{radford2016} showed that a GAN discriminator learns a hierarchical image representation that is useful for object classification.  
Such results suggest that combining an unsupervised GAN objective with a supervised classification objective would produce a simple and effective semi-supervised learning method. 
This approach, denoted by SSL-GAN, is illustrated in \fig{semisup}(b). The discriminator network receives a gradient from the real/fake loss for every real and generated image.
The discriminator also receives a gradient from the classification loss on the subset of (real) images for which labels are available. 

Generative adversarial networks have shown impressive performance on many diverse datasets. However, samples are most coherent when the set of images the network is trained on comes from a limited domain (eg. churches or faces).
Additionally, it is difficult to train GANs on very large images.
Both these issues suggest semi-supervised learning with vanilla GANs may not scale well to datasets of large diverse images. 
Rather than determining if a full image is real or fake, context conditional GANs address a different task: determining if {\it a part of an image} is real or fake {\it given the surrounding context}.

Formally, let $\mathcal{X_L} = \{(\mathbf{x}^1, y^1), ..., (\mathbf{x}^n, y^n)\}$ denote a dataset of labeled images.
Let $D_c(x)$ denote the output of the classifier head on the discriminator (see \fig{semisup}(c) for details). 
Then the semi-supervised CC-GAN objective is: 
\begin{align}
\label{eqn:sslccgan}
\min_G \max_D \hspace{2mm} & \mathbb{E}_{\mathbf{x} \sim \mathcal{X}}[\log D(\mathbf{x})]  
+ \mathbb{E}_{\mathbf{x} \sim \mathcal{X}, \mathbf{m} \sim \mathcal{M}} [\log(1 - D(\mathbf{x_I}))]
+ \lambda_c  \mathbb{E}_{\mathbf{x},y \sim \mathcal{X_L}} [\log(D_c(y | \mathbf{x}))]
\end{align}

The hyperparameter $\lambda_c$ balances the classification and adversarial losses. 
We only consider the CC-GAN in the semi-supervised setting and thus drop the SSL notation when referring to this model. 

\subsection{Model Architecture and Training Details}

\begin{figure}[t]
\centering
 \includegraphics[width=0.85\linewidth]{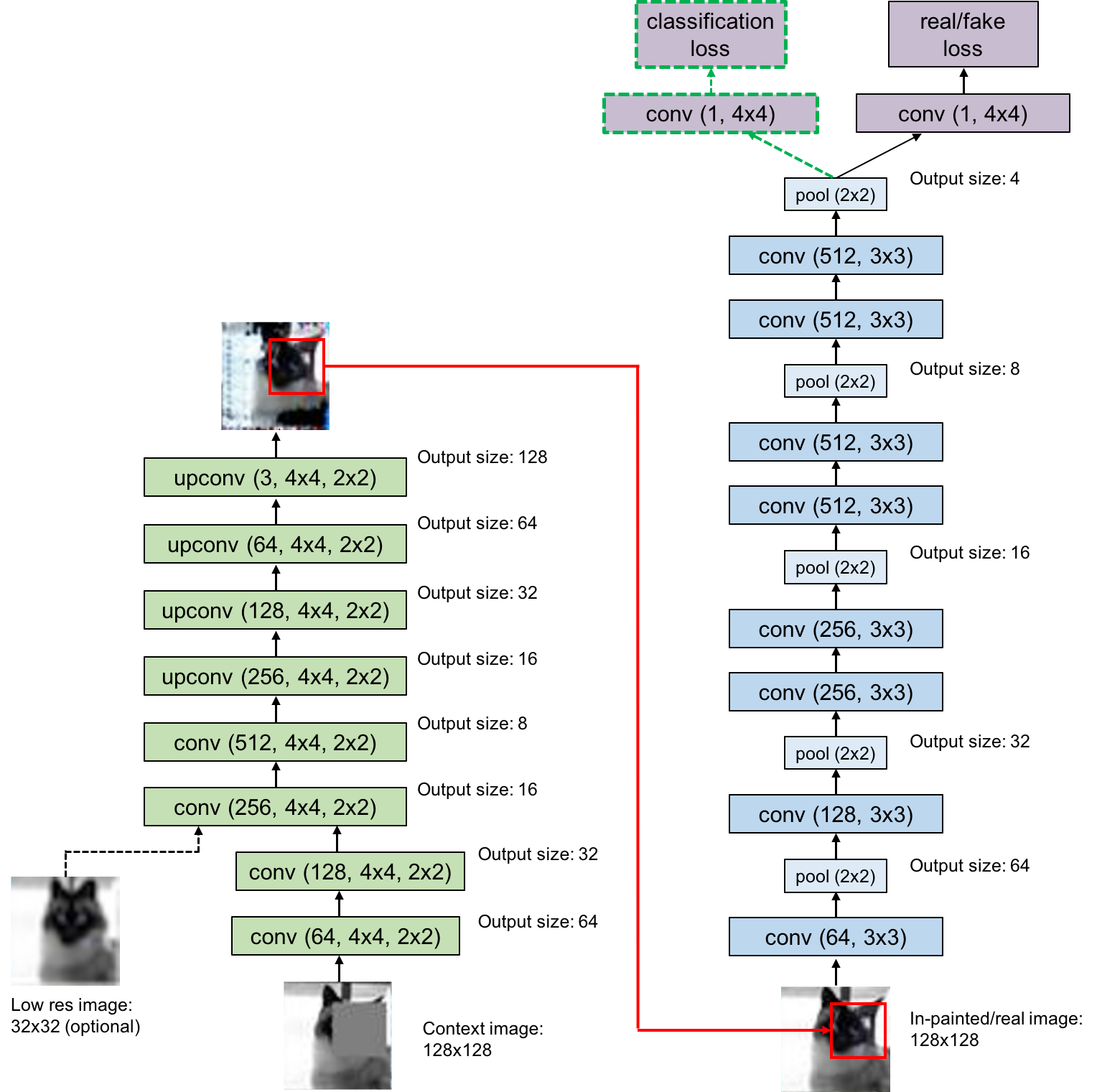} 
\caption{ Architecture of our context-conditional generative adversarial network (CC-GAN). \texttt{conv(64, 4x4, 2x2)} denotes a conv layer with 64 channels, 4x4 kernels and stride 2x2. Each convolution layer is followed by a spatial batch normalization and rectified linear layer. Dashed lines indicate optional pathways.
}
\vspace{-0.7em}
\label{fig:network}
\end{figure}

The architecture of our generative model, $G$, is inspired by the
generator architecture of the DCGAN \citep{radford2016}.  The model
consists of a sequence of convolutional layers with subsampling (but
no pooling) followed by a sequence of fractionally-strided
convolutional layers.  For the discriminator, $D$, we used the VGG-A
network \citep{vgg} without the fully connected layers (which we call
the VGG-A' architecture). Details of the
generator and discriminator are given in \fig{network}. The input to the generator is an image with a patch zeroed out. In preliminary
experiments we also tried passing in a separate mask to the generator
to make the missing area more explicit but found this did not effect
performance.

Even with the context conditioning it is difficult to generate large
image patches that look realistic, making it problematic to scale our
approach to high resolution images.  To address this, we propose
conditioning the generator on both the high resolution image with a
missing patch and a low resolution version of the whole image (with no
missing region).  
In this setting, the generator’s task becomes one of super-resolution on a portion of an image. 
However, the discriminator does not receive the low resolution image and thus is still faced with the same problem of determining if a given in-painting is viable or not. 
Where indicated, we used this approach in our PASCAL VOC 2007
experiments, with the original image being downsampled by a
factor of 4.  This provided enough information for the generator to
fill in larger holes but not so much that it made the task trivial.
This optional low resolution image is illustrated in
\fig{network}(left) with the dotted line.

We followed the training procedures of \cite{radford2016}.
We used the Adam optimizer \citep{kingma2015} in all our experiments with learning rate of 0.0002, momentum term $\beta_1$ of 0.5, and the remaining Adam hyperparameters set to their default values.
We set $\lambda_c = 1$ for all experiments.

\vspace{1em}
\section{Experiments}

\begin{table}[!t]
\centering
\small
\begin{tabularx}{0.7\textwidth}{l|l}
\bf{Method} & \bf{Accuracy}\\
\hline
\hline
Multi-task Bayesian Optimization \citep{swersky2013} & 70.10 $\pm$ 0.6 \\
Exemplar CNN \citep{dosovitskiy2014b} & 75.40 $\pm$ 0.3 \\
Stacked What-Where Autoencoder \citep{zhao2016}& 74.33 \\
\hline
Supervised VGG-A'  & 61.19 $\pm$ 1.1\\
SSL-GAN  & 73.81 $\pm$ 0.5\\
CC-GAN& 75.67 $\pm$ 0.5 \\
CC-GAN$^2$& \bf{77.79} $\pm$ 0.8 \\
\end{tabularx}
\vspace{2mm}
\caption{Comparison of CC-GAN and other published results on STL-10.}
\label{tab:stl}
\end{table}

\subsection{STL-10 classification}

STL-10 is a dataset of 96$\times$96 color images with a 1:100 ratio of labeled to unlabeled examples, making it an ideal fit for our semi-supervised learning framework. 
The training set consists of 5000 labeled images, mapped to 10 pre-defined folds of 1000 images each, and 100,000 unlabeled images.
The labeled images belong to 10 classes and were extracted from the ImageNet dataset and the unlabeled images come from a broader distribution of classes. 
We follow the standard testing protocol and train 10 different models on each of the 10 predefined folds of data. 
We then evaluate classification accuracy of each model on the test set and report the mean and standard deviation. 

We trained our CC-GAN and CC-GAN$^2$ models on 64$\times$64 crops of the 96$\times$96
image. The hole was 32$\times$32 pixels and the location of the hole
varied randomly (see \fig{ccgan_samples}(top)).  
We trained for 100 epochs and then 
fine-tuned the discriminator on the 96x96
labeled images, stopping when training accuracy reached 100\%. 
As shown in \tab{stl}, the CC-GAN model performs comparably to current state of the art \citep{dosovitskiy2014b} and the CC-GAN$^2$ model improves upon it.

We also trained two baseline models in an attempt to tease apart the
contributions of adversarial training and context conditional
adversarial training.  The first is a purely supervised training of
the VGG-A' model (the same architecture as the discriminator in the
CC-GAN framework). This was trained using a dropout of 0.5 on the
final layer and weight decay of 0.001.  The performance of this model is significantly worse than the CC-GAN model.

We also trained a semi-supervised GAN (SSL-GAN, see Fig.~1(b)) on STL-10. This
consisted of the same discriminator as the CC-GAN (VGG-A'
architecture) and generator from the DCGAN model
\citep{radford2016}. The training setup in this case is identical to
the CC-GAN model. The SSL-GAN performs almost as well as the CC-GAN,
confirming our hypothesis that the GAN objective is a useful unsupervised criterion.

\subsection{PASCAL VOC classification}
In order to compare against other methods that utilize spatial context
we ran the CC-GAN model on PASCAL VOC 2007 dataset.  This dataset
consists of natural images coming from 20 classes.  The dataset
contains a large amount of variability with objects varying in size,
pose, and position.  The training and validation sets combined
contain 5,011 images, and the test set contains 4,952 images. The
standard measure of performance is mean average precision (mAP).

We trained each model on the combined training and validation set for
$\sim$5000 epochs and evaluated on the test set
once\footnote{Hyperparameters were determined by initially training on
  the training set alone and measuring performance on the validation
  set.}. Following \cite{pathak2016}, we train using
random cropping, and then evaluate using the average prediction from
10 random crops.

Our best performing model was trained on images of resolution
128$\times$128 with a hole size of 64$\times$64 and a low resolution
input of size 32$\times$32.  \tab{pascal} compares our CC-GAN method
to other feature learning approaches on the PASCAL test set. It
outperforms them, beating the current state of the art \citep{wang2015} by 3.8\%. It is
important to note that our feature extractor is the VGG-A' model which
is larger than the AlexNet architecture \citep{krizhevsky2012} used by
other approaches in \tab{pascal}.  However, purely supervised training
of the two models reveals that VGG-A' is less than 2\% better than
AlexNet. Furthermore, our model outperforms the supervised VGG-A'
baseline by a 7\% margin (62.2\% vs. 55.2\%).  This suggests that our
gains stem from the CC-GAN method rather than the use of a better
architecture.

\tab{pascal_detail} shows the effect of training on different
resolutions.  The CC-GAN improves over the baseline CNN consistently
regardless of image size.  We found that conditioning on the low
resolution image began to help when the hole size was largest
(64$\times$64).  We hypothesize that the low resolution conditioning
would be more important for larger images, potentially allowing the
method to scale to larger image sizes than we explored in this work.

\begin{table}[t]
\centering
\small
\begin{tabularx}{0.7\textwidth}{l|c}
\bf{Method} & \bf{mAP}\\
\hline
\hline
Supervised AlexNet & 53.3 \%\\
Visual tracking from video \citep{wang2015} & 58.4\% \\
Context prediction \citep{doersch2015} & 55.3\%\\
Context encoders \citep{pathak2016} & 56.5\% \\
\hline
Supervised VGG-A' & 55.2\%\\
CC-GAN  & 62.2\%\\
CC-GAN$^2$  & {\bf 62.7}\%\\
\end{tabularx}
\vspace{2mm}
\caption{Comparison of CC-GAN and other methods (as reported by \cite{pathak2016}) on PASCAL VOC 2007.}
\label{tab:pascal}
\end{table}

\begin{table}[!t]
\centering
\small
\begin{tabularx}{0.8\textwidth}{l|c|c|c|c}
\bf{Method} & {\bf Image size} & {\bf Hole size} & {\bf Low res size} & \bf{mAP}\\
\hline
\hline
Supervised VGG-A' & 64$\times$64 & - & - & 52.97\%\\
CC-GAN & 64$\times$64 & 32$\times$32 & - & 56.79\%\\
\hline
Supervised VGG-A' & 96$\times$96 & - & - & 55.22\%\\
CC-GAN & 96$\times$96 & 48$\times$48 & - & 60.38\%\\
CC-GAN & 96$\times$96 & 48$\times$48 & 24$\times$24 & 60.98\%\\
\hline
Supervised VGG-A' & 128$\times$128 & - & - & 55.2\%\\
CC-GAN & 128$\times$128 & 64$\times$64 & - & 61.3\%\\
CC-GAN & 128$\times$128 & 64$\times$64 & 32$\times$32  & 62.2\%\\
\end{tabularx}
\vspace{2mm}
\caption{Comparison of different CC-GAN variants on PASCAL VOC 2007.}
\label{tab:pascal_detail}
\end{table}

\subsection{Inpainting}
\begin{figure}[t!]
\centering
  \includegraphics[width=1\linewidth]{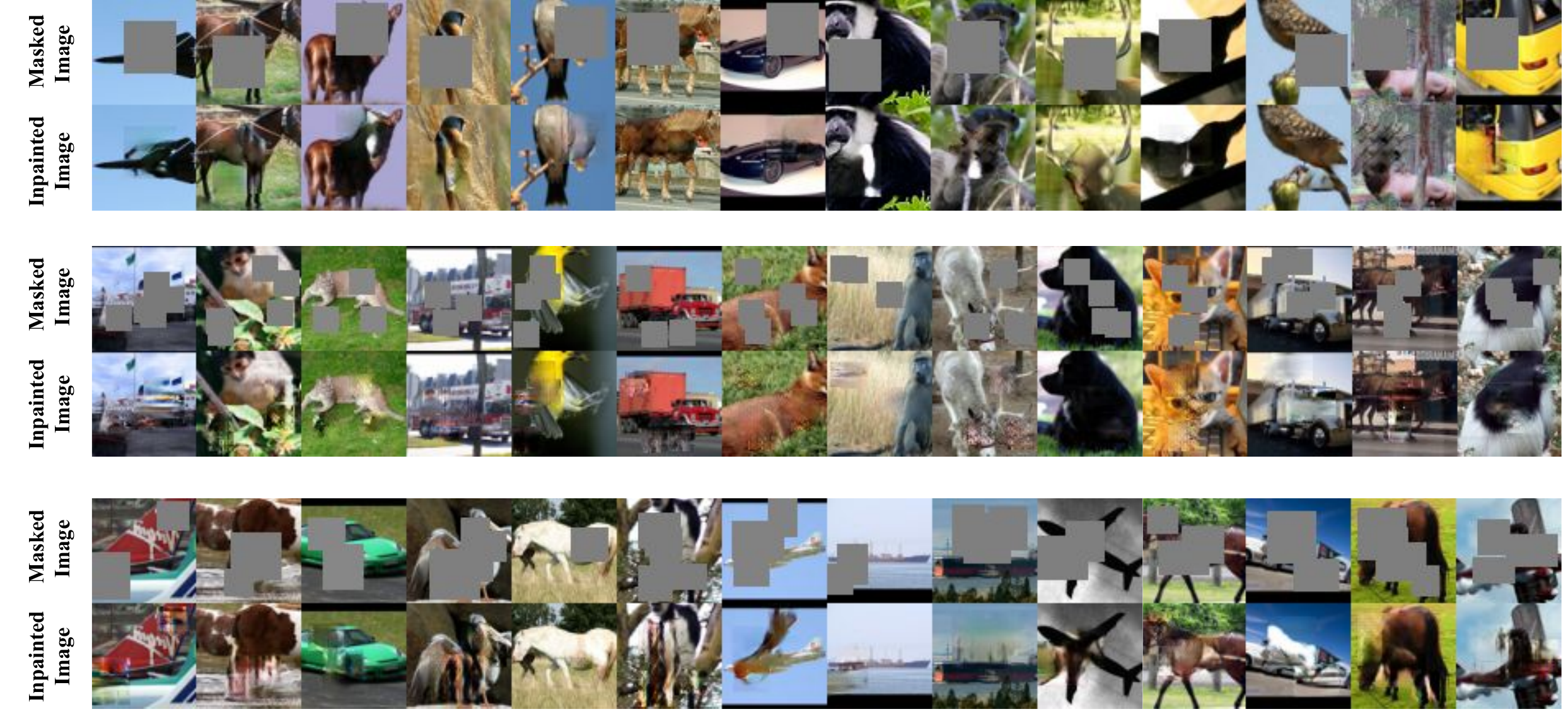} 
\caption{STL-10 in-painting with CC-GAN training and varying methods of dropping out the image. }
\label{fig:ccgan_samples}
\vspace{2mm}
\end{figure}

We now show some sample in-paintings produced by our CC-GAN
generators.  In our semi-supervised learning experiments on STL-10 we
remove a single fixed size hole from the image.  The top row of
\fig{ccgan_samples} shows in-paintings produced by this model.  We can
also explored different masking schemes as illustrated in the
remaining rows of \fig{ccgan_samples} (however these did not improve
classification results).  In all cases we see that
training the generator with the adversarial loss produces sharp
semantically plausible in-painting results.

\begin{figure}[t!]
\centering
  \includegraphics[width=1\linewidth]{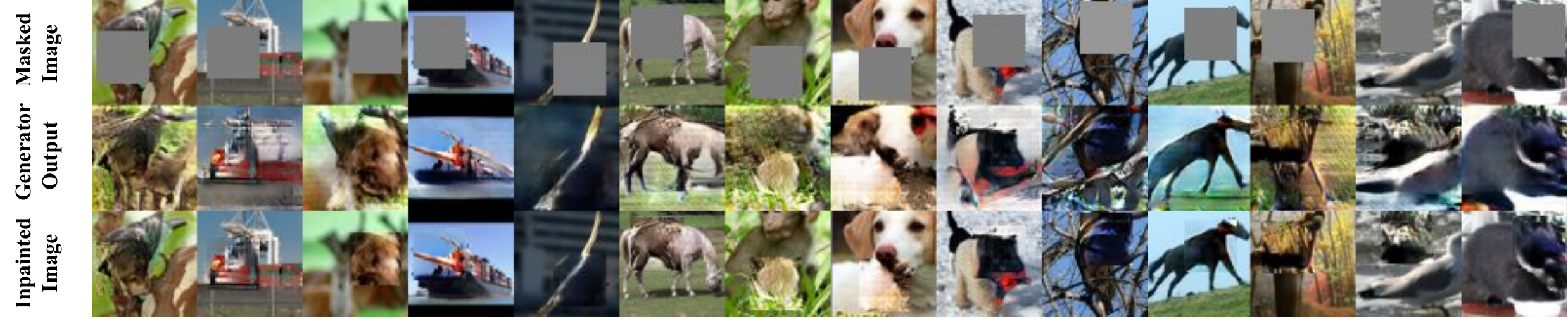} 
\caption{STL-10 in-painting with combined CC-GAN$^2$ training. }
\vspace{2mm}
\label{fig:ccgan_comb_samples}
\end{figure}

\fig{ccgan_comb_samples} shows generated images and in-painted images
from a model trained with the CC-GAN$^2$ criterion.  The output of a
CC-GAN generator tends to be corrupted outside the patch used to
in-paint the image (since gradients only flow back to the missing patch). However, in the CC-GAN$^2$ model, we see that both the
in-painted image and the generated image are coherent and semantically
consistent with the masked input image.

\begin{figure}[t!]
\centering
  \includegraphics[width=1\linewidth]{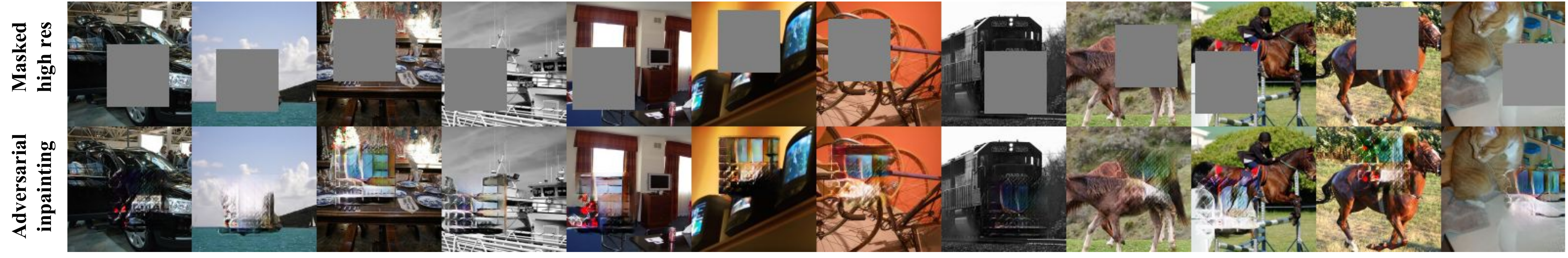} 
\caption{PASCAL in-painting with CC-GAN.}
\vspace{2mm}
\label{fig:pascal_samples_nolowres}
\end{figure}

\begin{figure}[t!]
\centering
  \includegraphics[width=1\linewidth]{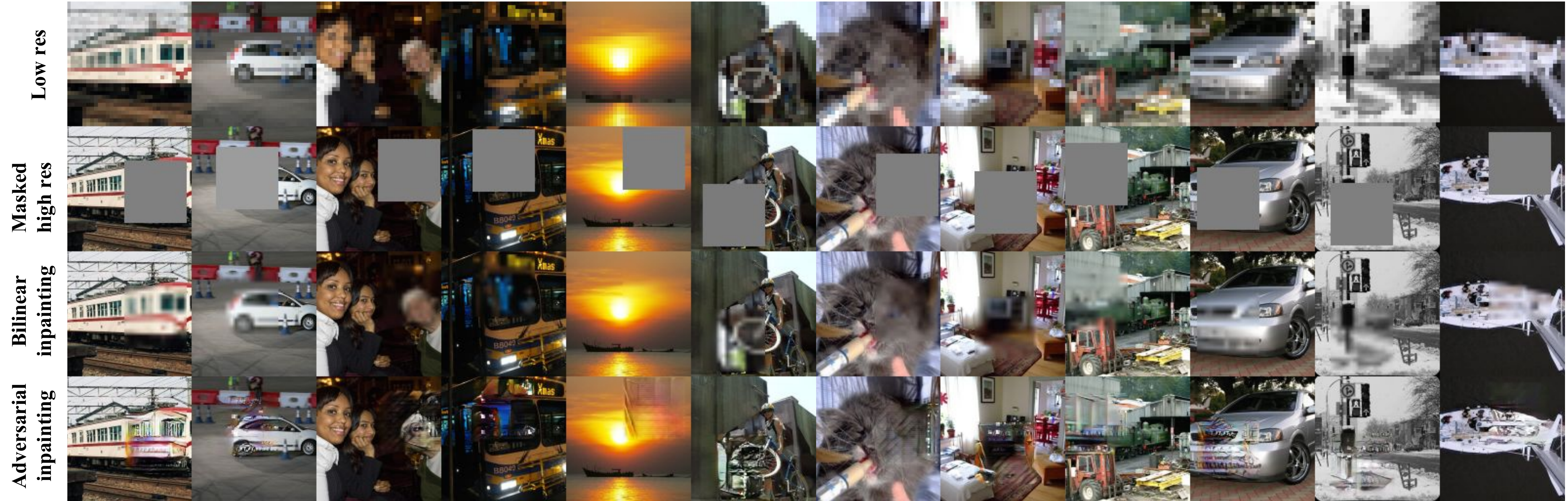} 
\caption{PASCAL in-painting with CC-GAN conditioned on low resolution image. Top two rows show input to generator. Third row shows inpainting my bilinear upsampling. Bottom row shows inpainted image by generator. }
\label{fig:pascal_samples}
\end{figure}

\fig{pascal_samples_nolowres} shows in-painted images from a generator
trained on 128$\times$128 PASCAL images.  \fig{pascal_samples} shows
the effect of adding a low resolution (32$\times$32) image as input to
the generator.  For comparison we also show the result of in-painting
by filling in with a bi-linearly upsampled image.  Here we see the
generator produces high-frequency structure rather than simply learning to copy the low resolution patch.

\vspace{10mm}

\section{Discussion}
We have presented a simple semi-supervised learning framework based on
in-painting with an adversarial loss.  The generator in our CC-GAN
model is capable of producing semantically meaningful in-paintings and
the discriminator performs comparable to or better than existing
semi-supervised methods on two classification benchmarks.

Since discrimination of real/fake in-paintings is more closely related
to the target task of object classification than extracting a feature
representation suitable for in-filling, it is not surprising that we
are able to exceed the performance of \cite{pathak2016}
on PASCAL classification.  Furthermore, since our model operates on
images half the resolution as those used by other approaches
(128$\times$128 vs. 224$\times$244), there is potential for further
gains if improvements in the generator resolution can be made. Our models and code are available at \url{https://github.com/edenton/cc-gan}.

\noindent {\bf Acknowledgements:} Emily Denton is supported by a
Google Fellowship. Rob Fergus is grateful for the support of CIFAR.

\small
\bibliography{bibliography} 
\bibliographystyle{iclr2017_conference}

\end{document}